\documentclass{article}
\usepackage{amsmath,epsfig}
\usepackage[preprint]{spconf}

\usepackage{amssymb}
\usepackage{multirow}
\usepackage{subfigure}

\let\OLDthebibliography\thebibliography
\renewcommand\thebibliography[1]{
  \OLDthebibliography{#1}
  \setlength{\parskip}{0pt}
  \setlength{\itemsep}{0pt plus 0.3ex}
}

\begin{document}\sloppy

\title{Modulating Localization and Classification for Harmonized Object Detection}
%
\name{Taiheng Zhang$^{1\dagger}$, Qiaoyong Zhong$^2$, Shiliang Pu$^{2\star}$, Di Xie$^2$}
\address{$^1$Zhejiang University\quad$^2$Hikvision Research Institute\\
thzhang@zju.edu.cn, \{zhongqiaoyong,pushiliang.hri,xiedi\}@hikvision.com}

\maketitle

\let\thefootnote\relax\footnotetext{$^\dagger$Work done as an intern at Hikvision Research Institute.}
\let\thefootnote\relax\footnotetext{$^\star$Corresponding author.}

\begin{abstract}
  Object detection involves two sub-tasks, i.e. localizing objects in an image and classifying them into various categories. For existing CNN-based detectors, we notice the widespread divergence between localization and classification, which leads to degradation in performance. In this work, we propose a mutual learning framework to modulate the two tasks. In particular, the two tasks are forced to learn from each other with a novel mutual labeling strategy. Besides, we introduce a simple yet effective IoU rescoring scheme, which further reduces the divergence. Moreover, we define a Spearman rank correlation-based metric to quantify the divergence, which correlates well with the detection performance. The proposed approach is general-purpose and can be easily injected into existing detectors such as FCOS and RetinaNet. We achieve a significant performance gain over the baseline detectors on the COCO dataset.
\end{abstract}
\begin{keywords}
Object Detection, Mutual Learning
\end{keywords}
\section{Introduction}

Object detection is a fundamental task in computer vision. It serves as a key component for a broad set of downstream vision applications, such as instance segmentation~\cite{he2017mask} and human pose estimation~\cite{papandreou2017towards}. Over the past few years we have witnessed the success of convolutional neural networks (CNNs) for object detection. CNN-based detectors have been evolving quickly, resulting in various distinct frameworks, such as anchor-based~\cite{liu2016ssd,ren2017faster,lin2017focal} and anchor-free~\cite{yu2016unitbox,tian2019fcos} methods.

The task of object detection involves two sub-tasks, i.e. localizing objects in an image and classifying them into various categories. Accordingly, most existing CNN-based detectors employ two network branches. The localization branch estimates the location and scale of objects and the classification branch predicts the confidence of each class that an object may belong to. Given dense bounding box predictions by CNN, non-maximum suppression (NMS) is commonly applied to remove redundant boxes as a post-processing step. In NMS, the classification confidence is used as a measure of quality of the boxes. The boxes are ranked by their confidence values. Then the boxes with high confidence are selected, while those with low confidence are suppressed.

\begin{figure}[t]
\centering
\subfigure[]{
\label{divergencea}
\includegraphics[width=0.6\linewidth]{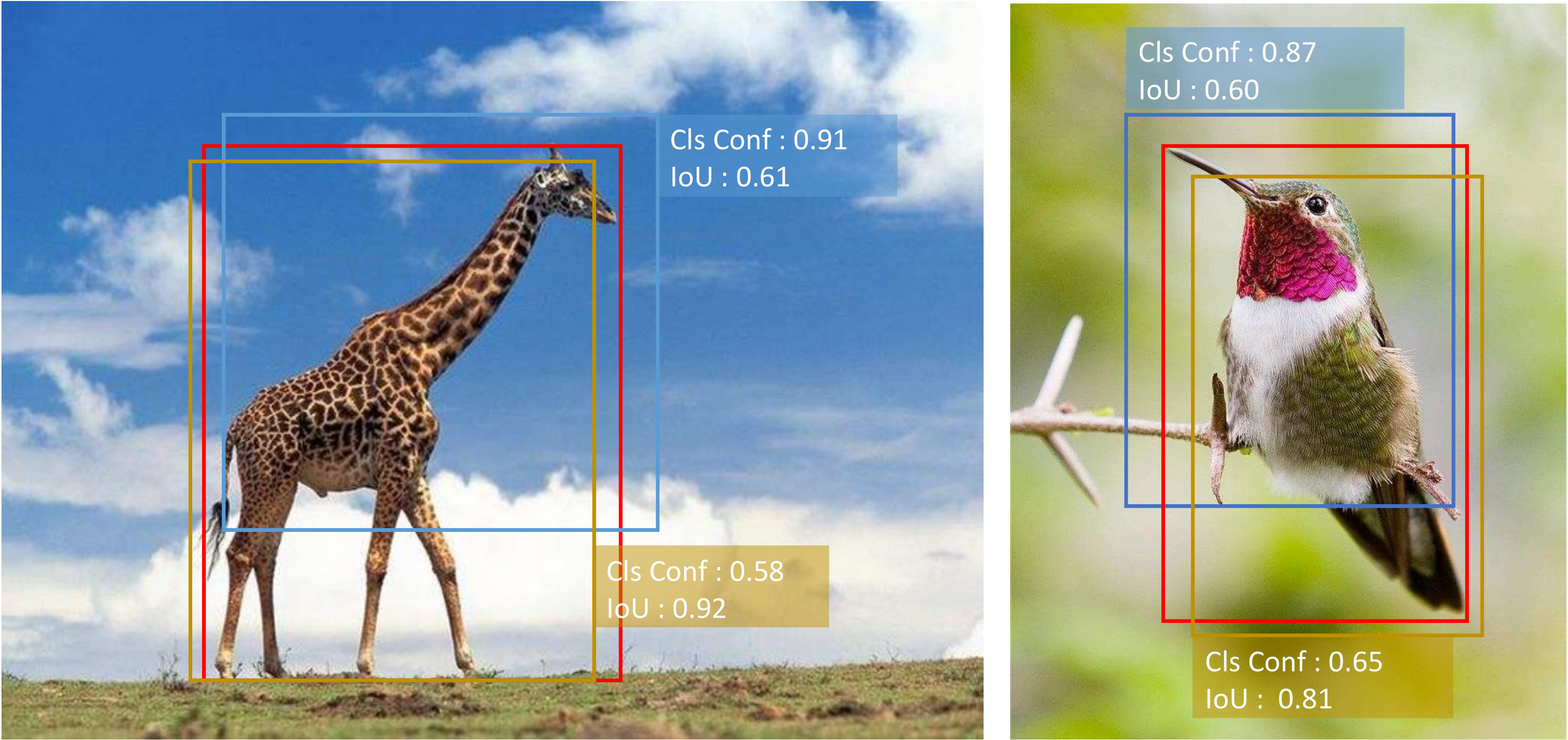}
}\hspace{1em}%
\subfigure[]{
\label{divergenceb}
\includegraphics[width=0.3\linewidth]{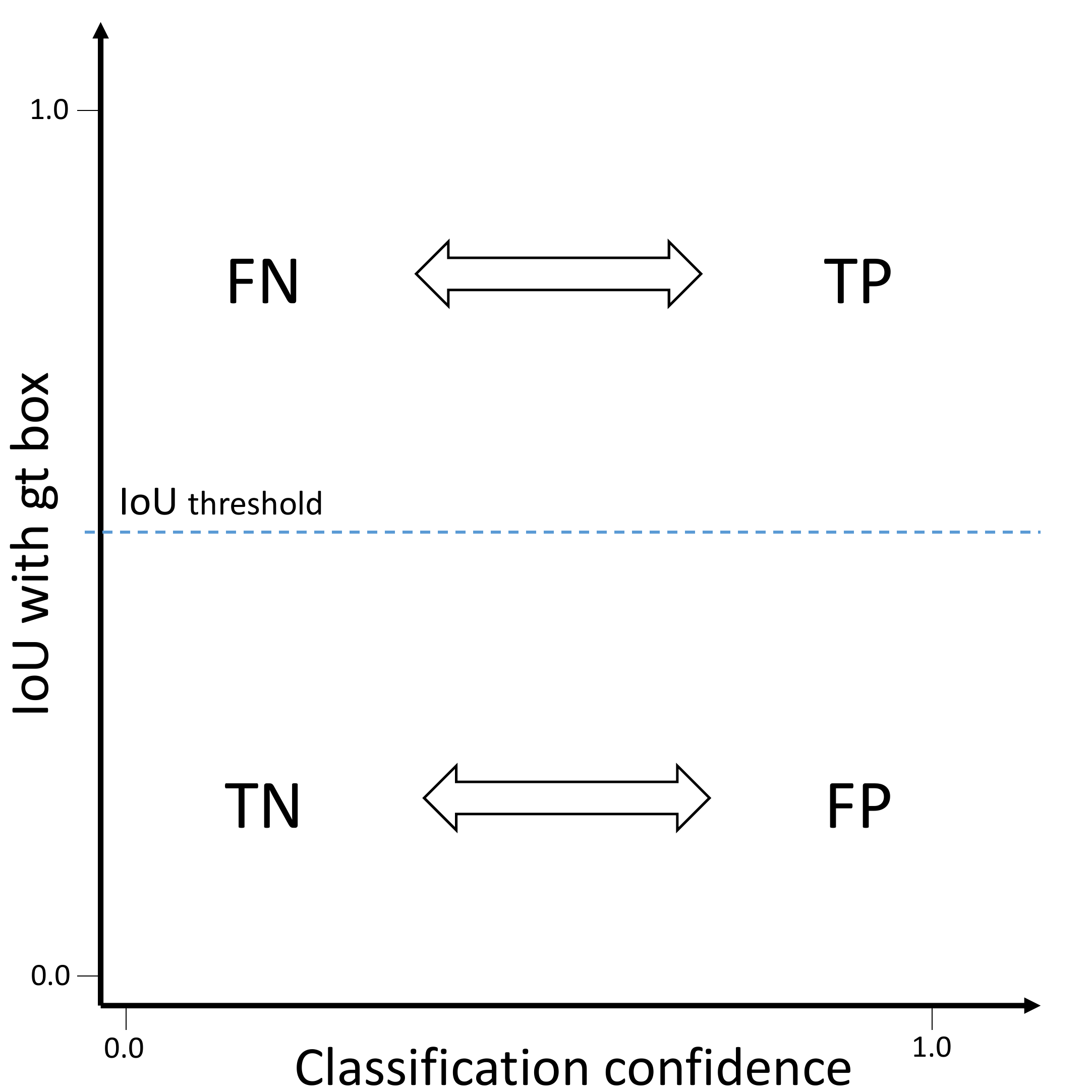}
}
\caption{The divergence between classification and localization. (a) Two examples of divergence. The red box is the ground truth and the blue and yellow boxes are the detected boxes before NMS. (b) Divergence can cause missing detections (FN) and false alarms (FP).}
\label{fig:mutual learning}
\end{figure}

A good detection for an object requires both high classification accuracy and localization quality. The classification accuracy is measured by the predicted confidence of the ground-truth class. The localization quality can be measured by its intersection-over-union (IoU) with the matched ground-truth box~\cite{jiang2018acquisition}. The ideal circumstance for NMS to hit good detections and suppress bad detections is that the classification confidence and localization quality are positively correlated. Although the classification and localization tasks can be learned jointly in a multi-task manner in existing CNN-based detectors, their training targets are set independently. In reality, the ideal condition can be hardly satisfied. A common phenomenon is schematically shown in Fig.~\ref{divergencea}, in which a box with high classification confidence is less accurately localized (with lower IoU) than a box with low confidence. The divergence comes from the fact that classification and localization may prefer different prior reference points (e.g. anchors for anchor-based detectors). For example, the reference point near the head of the bird in Fig.~\ref{divergencea} contains the most discriminative feature to distinguish the bird from other categories, while it is easier for the reference point near the body of the bird to accurately estimate the bounding box.

Due to the divergence, the classification confidence of a candidate box is not able to fully represent its quality. Thus NMS suffers from selecting boxes of higher confidence and lower IoU and suppressing boxes of lower confidence and higher IoU. As illustrated in Fig.~\ref{divergenceb}, this issue leads to both false alarms and missing detections. As a common setting, a detection is considered correct if its IoU with the ground-truth box is above a threshold (e.g. 0.5). For boxes with IoU above the threshold, those with high confidence are true positives, while those with low confidence are prone to cause missing detections. For boxes with IoU below the threshold, those with high confidence are prone to cause false alarms.

In this work, we aim to modulate the classification and localization tasks such that the problem of divergence gets alleviated. In particular, we introduce the concept of mutual learning~\cite{zhang2018deep} in the context of object detection. A novel mutual labeling (ML) strategy is proposed, where the training target of a box for classification is determined by its localization quality and vice versa. In this way, the two tasks are forced to learn from each other, and the divergence gets reduced gradually during the training procedure. Moreover, inspired by IoU-Net~\cite{jiang2018acquisition} and FCOS~\cite{tian2019fcos}, we propose a simple IoU rescoring scheme (IUR), which further reduces the divergence. The overall framework, named modulating localization and classification (MLC) is general-purpose and applies to detectors of various types, such as FCOS~\cite{tian2019fcos} (anchor-free) and RetinaNet~\cite{lin2017focal} (anchor-based). MLC leads to harmonized object detection and improves the baseline detectors significantly on the commonly used COCO dataset~\cite{lin2014microsoft}.

To systematically investigate the influence of divergence, we propose a Spearman rank correlation-based metric to quantify the degree of divergence. Experiments show that this metric correlates well with the final detection performance.

Our major contributions are summarized as follows.
\begin{itemize}
  \item We propose a novel mutual labeling strategy to address the divergence between localization and classification in object detection.
  \item We introduce a simple yet effective IoU rescoring scheme to further reduce the divergence.
  \item We define a Spearman rank correlation-based metric to measure the divergence, which is valuable for quantitative analysis of the problem.
  \item The proposed approach applies to detectors of various types, and improves the baseline detectors significantly.
\end{itemize}

\begin{figure}[t]
\centering
\includegraphics[width=0.8\linewidth]{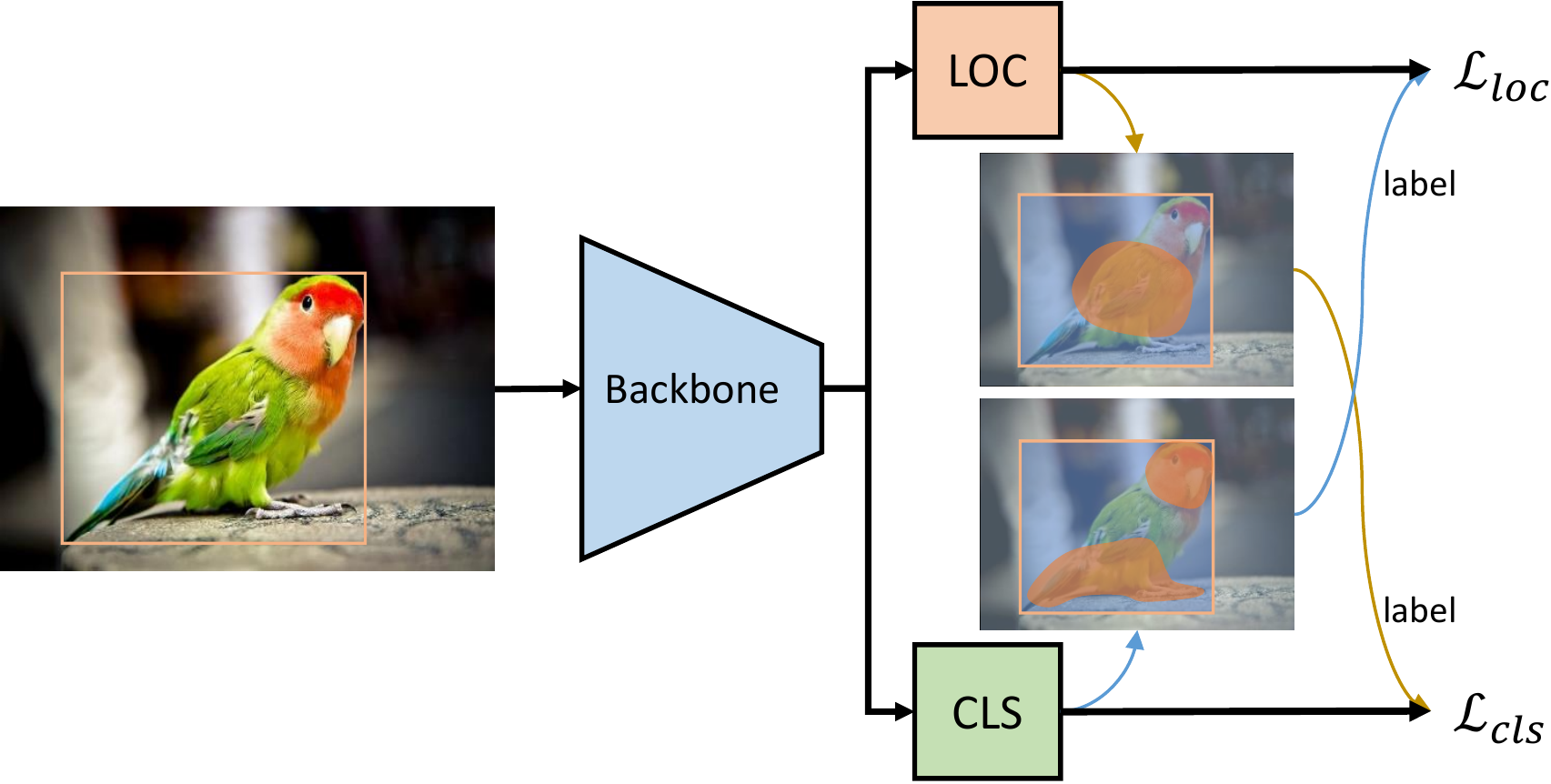}
\caption{The mutual labeling strategy. The orange pixels overlaid on the parrot represent the mutually labeled positive samples, and the blue pixels are negative samples.}
\label{fig:mut-lab}
\end{figure}

\section{Related Work}
\label{sec:related-work}

\noindent \textbf{CNN-based Object Detectors.} In terms of the number of processing stages, existing object detection methods can be grouped into one-stage and multi-stage detectors. One-stage detectors, such as YOLO~\cite{redmon2016you,redmon2017yolo9000,redmon2018yolov3}, SSD~\cite{liu2016ssd} and RetinaNet~\cite{lin2017focal} predict the detections using a single network. Two-stage detectors, such as the R-CNN series~\cite{girshick2014rich,girshick2015fast,ren2017faster,zhong2020} decompose the task into two sequential steps, i.e. the region proposal step and the region classification step.
To ease regression of the target bounding boxes, the design of anchor boxes was first introduced in~\cite{ren2017faster} and widely adopted in subsequent works~\cite{liu2016ssd,redmon2017yolo9000,lin2017focal,lin2017feature}. On the other hand, there are other works attempting to remove the dependency of anchors by predicting the absolute location and scale of target objects~\cite{huang2015densebox,redmon2016you,yu2016unitbox,tian2019fcos}. Although these detectors are distinct from each other, most of them involve the object classification and localization tasks. By modulating the learning procedure of the two tasks, our approach is general-purpose and beneficial to most existing detectors.

\noindent \textbf{Divergence between Classification and Localization.} Few works have investigated the divergence problem in the literature. The most related work is IoU-Net~\cite{jiang2018acquisition}, which attempts to accurately predict the IoUs of the detected boxes to their matched ground-truth boxes. Then the learned IoU is exploited to guide NMS such that accurately localized boxes are more likely to survive. In our work, we aim to reach a consensus between the two tasks by mutually mining the abundant positive samples, which addresses the divergence problem more directly and naturally. Besides, we introduce a simple IoU rescoring scheme, which is shown effective without the requirement of highly accurate IoU prediction.

\noindent \textbf{Mutual Learning between Networks.} Zhang et al.~\cite{zhang2018deep} presented a deep mutual learning strategy where an ensemble of small networks learn collaboratively against the same task and teach each other throughout the training procedure. We introduce this concept in the context of object detection and setup a mutual learning framework between the classification and localization tasks. In contrast to~\cite{zhang2018deep}, our approach involves mutual learning of two different tasks.

\noindent \textbf{Leveraging Ignored Samples.} When assigning the target label of the priors (e.g. anchor boxes), most detectors ignore the samples of ambiguity during training. For example, in RetinaNet~\cite{lin2017focal}, an anchor is labeled as positive if its IoU with the ground-truth box is greater than 0.5, negative if IoU is less than 0.4, and ignored otherwise. We argue that the ignored samples are valuable for learning. In this work, we offer a way to exploit the ignored samples via mutual labeling.

\noindent \textbf{Adaptive Anchor Assignment.}
Recently, some works attempted to assign the labels of anchors adaptively. ATSS~\cite{zhang2020bridging} sets a dynamic IoU threshold according to the statistics of an object. MAL~\cite{ke2020multiple} selects positive anchors by combining their classification and localization scores. PAA~\cite{kim2020probabilistic} uses Gaussian Mixture Model to separate positive and negative anchors based on a carefully designed anchor score. AutoAssign~\cite{zhu2020autoassign} introduces a confidence weighting module for adaptive assignment. Our new contribution is the mutual learning strategy between the classification and localization tasks.

\section{Our Approach}

\subsection{Mutual Labeling for Mutual Learning}
\label{sect:mut-lab}

Let us first recap the common form of loss function used in a typical detector. By matching the priors to their best matched ground truth, they are labeled as positive or negative. Then the loss function can be defined as
\begin{equation}
  \label{eq-loss-det}
  \mathcal L_{det}=\mathcal L_{cls}(A_+ \cup A_-) + \mathcal L_{loc}(A_+).
\end{equation}
It consists of the classification loss $\mathcal L_{cls}$ imposed on both positive samples $A_+$ and negative samples $A_-$, and the localization loss $\mathcal L_{loc}$ imposed on positive samples only.
\begin{equation}
  \label{eq-loss-cls}
  \mathcal L_{cls}(A)=\frac{1}{|A|}\sum_{j \in A} W_j^{cls}\ell_{cls}(c_{j},l_j^{gt})
\end{equation}
\begin{equation}
  \label{eq-loss-loc}
  \mathcal L_{loc}(A)=\frac{1}{|A|}\sum_{j \in A} W_j^{loc}\ell_{loc}(b_{j},b_j^{gt})
\end{equation}
where $\ell_{cls}$ and $\ell_{loc}$ are the loss upon a single sample $j$. A common choice is to use the cross entropy loss for $\ell_{cls}$ and smooth L1 regression loss for $\ell_{loc}$. $W_j$ is the loss weight of sample $j$ which is usually set to 1. $c_{j}$ and $b_{j}$ are the predicted confidence and the estimated bounding box for $j$. $l_{j}^{gt}$ and $b_{j}^{gt}$ are the ground-truth class label and target box assigned to $j$.

To address the divergence between localization and classification, we propose a novel mutual labeling (ML) strategy. As illustrated in Fig.~\ref{fig:mut-lab}, we set the training targets of the classification and localization tasks mutually in a crossing manner. For each sample, its target label of localization is decided by thresholding its classification quality measured by the score of the correct class. Analogously, the target label of classification is decided by thresholding the localization quality measured by IoU of the estimated bounding box with the matched ground-truth box. Note that the target labels here refer to the grouping of positive and negative samples, not the specific training targets of the two tasks (one-hot class label for classification and ground-truth box for localization). With the mutual labeling strategy, the model is forced to automatically discover the consensus samples where the classification and localization tasks agree to each other.

We use Otsu's method~\cite{otsu1979a} to compute the threshold values in mutual labeling. The thresholding is performed in a per-object manner. Specifically, all candidate positive samples matched to a single object form a group and are divided into positive and negative samples. It can be formulated as
\begin{equation}
  \label{eq:J-plus-cls}
  J_{+}^{cls,k}=\left\{j|I_j^{k}>Otsu(I^{k}),j\in J^{k} \right\}
\end{equation}
\begin{equation}
  \label{eq:J-minus-cls}
  J_{-}^{cls,k}=\left\{j|I_j^{k}\le Otsu(I^{k}),j\in J^{k} \right\}
\end{equation}
\begin{equation}
  \label{eq:J-plus-loc}
  J_{+}^{loc,k}=\left\{j|S_j^{k}>Otsu(S^{k}),j\in J^{k} \right\}
\end{equation}
where $k$ denotes an object and $J^{k}$ denotes the set of candidate positive samples matched to $k$. $S^{k}$ is the set of confidence values of samples in $J^{k}$ and $I^{k}$ is the set of IoU values of samples in $J^{k}$. By computing the union of positive and negative samples $J_{+}^{cls,k}$, $J_{-}^{cls,k}$ and $J_{+}^{loc,k}$ over all objects $k$, we get the full sets of mutually labeled samples $A_{+}^{cls}$, $A_{-}^{cls}$ and $A_{+}^{loc}$.

To strengthen the mutual learning process, we leverage all samples that are likely to be positive for training. For instance, in RetinaNet~\cite{lin2017focal} all of the ignored samples whose IoUs with ground-truth boxes range from 0.4 and 0.5 are treated as candidate positive samples. In FCOS~\cite{tian2019fcos} all samples located inside the ground-truth boxes are taken into account. By leveraging the ignored samples, we aim to mine good detections from a large set of candidate samples. To stabilize training, we reduce the loss weight of the originally ignored samples according to the margin to the threshold by
\begin{equation}
  \label{eq:W-cls}
W_j^{cls}=
\begin{cases}
|I_j-\tau^{loc}|^{\alpha},& \text{if $j$ is an ignored sample}\\
1,& \text{otherwise}
\end{cases}
\end{equation}
\begin{equation}
  \label{eq:W-loc}
W_j^{loc}=
\begin{cases}
|S_j-\tau^{cls}|^{\alpha},& \text{if $j$ is an ignored sample}\\
1,&\text{otherwise}
\end{cases}
\end{equation}
where $\alpha$ is a parameter to control the contribution of the ignored samples for training. $\tau^{loc}$ and $\tau^{cls}$ are the per-object thresholds computed using Otsu's method.

By mutual labeling, we divide the candidate positive samples into real positive and negative, which changes dynamically throughout the training procedure. Note that we essentially redefine the assignment of positive and negative labels to candidate positive samples. Thus the form of loss function is the same as the baseline detectors, which is updated as
\begin{equation}
\mathcal L_{det}^{ml}=\mathcal L_{cls}(A_{+}^{cls}\cup A_{-}^{cls}\cup A_{-}) + \mathcal L_{loc}(A_{+}^{loc})
  \label{eq:loss-ml}
\end{equation}

\subsection{IoU Rescoring}

To further reduce the divergence, we introduce a simple IoU rescoring (IUR) scheme. Inspired by IoU-Net~\cite{jiang2018acquisition} and the centerness in FCOS~\cite{tian2019fcos}, we learn to predict the IoU between the estimated bounding box and the ground-truth box by appending a single convolutional layer. During inference, the product of the predicted IoU and classification confidence is treated as a more reliable quality measure of the detections. The standard NMS is employed, where the confidence is replaced by the integrated quality score for ranking of the detections.

Since the IoU prediction task is more related to the localization task than the classification task, we append the IoU prediction layer to the localization branch rather than the classification branch as the centerness prediction layer in~\cite{tian2019fcos}. The loss function of IUR is defined as
\begin{equation}
  \label{eq:loss-iur}
\mathcal L_{iur}(A)=\frac{1}{|A|}\sum_{j \in A} \ell_{iur}(P_{j},I_{j})
\end{equation}
where $P_j$ denotes the predicted IoU of sample $j$ and $\ell_{iur}$ is set to the MSE loss in our experiments.

By combining mutual labeling and IUR, we are able to modulate the localization and classification (MLC) tasks for harmonized object detection. The total loss of MLC can be written as
\begin{equation}
  \label{eq:loss-mlc}
\mathcal L_{mlc}=\mathcal L_{det}^{ml}+\gamma \mathcal L_{iur}(A_{+}^{cls}\cup A_{+}^{loc})
\end{equation}
where the IUR loss is applied to the union of positive samples of both classification and localization tasks. $\gamma$ is the weight to balance the loss.

Although the divergence problem has been discussed in IoU-Net~\cite{jiang2018acquisition}, we make the following new contributions.
1) We aim to investigate the cause of divergence, i.e. the classification and localization tasks diverge from each other on deciding whether a prior box leads to a good detection or not. Accordingly we propose the mutual learning strategy, which works well without the necessity of introducing a new network branch.
2) Although the idea of IoU prediction is not new, we repurpose it to reduce the divergence by integrating it with the classification score. We intend to keep it light-weight so that it can easily complement the mutual learning strategy. It works well without complex architecture (a single convolutional layer versus an R-CNN-like head) and carefully designed sampling strategy and feature pooling as IoU-Net does.
3) Thanks to the ``non-invasive'' property of MLC, we can easily inject it into detectors of distinct architectures, such as FCOS and RetinaNet. While IoU-Net is preferably applied to the region classification stage of two-stage detectors only. Besides, MLC introduces marginal network parameters and extra computational cost during inference.

\begin{table}[t]
\small
\centering
\caption{Comparison of the FCOS$^-$ baseline, mutual learning with prediction alignment and mutual labeling in terms of the Spearman rank correlation and AP performance.}
\label{tab:comp-ml}
\begin{tabular}{l|c|c}
\hline
Method & Correlation & AP (\%)\\
\hline
FCOS$^{-}$ & 0.32 & 34.5 \\
FCOS$^{-}$ + Prediction Alignment & 0.42 & 35.3 \\
FCOS$^{-}$ + Mutual Labeling & \textbf{0.48} & \textbf{37.5}\\
\hline
\end{tabular}
\end{table}

\section{Experiments}

\begin{table}[t]
\small
\centering
\caption{Comparison of IUR with existing methods.}
\label{tab:comp-iur}
\begin{tabular}{l|c|c}
\hline
Method & Correlation & AP (\%) \\
\hline
FCOS$^{-}$ & 0.32 &34.5 \\
FCOS$^{-}$ + Centerness & 0.33 &36.9 \\
FCOS$^{-}$ + IUR & \textbf{0.50} & \textbf{37.3}\\
FCOS$^{-}$ + IoU-NMS & - & 32.6\\
\hline
\end{tabular}
\end{table}

To validate the effectiveness and versatility of our approach, we integrate it into three state-of-the-art detectors of various types, namely FCOS~\cite{tian2019fcos}, RetinaNet~\cite{lin2017focal} and RPN~\cite{ren2017faster} (for region proposal only). To compare our approach with the centerness prediction method in FCOS, we train an FCOS model without the centerness prediction as a reference (referred to as FCOS$^-$ henceforth). Our experiments are conducted on the MS COCO 2017~\cite{lin2014microsoft} dataset, which contains 118k images for training and 5k images for validation. All models are trained on the training set and evaluated on the validation set. The average mAP over IoU thresholds ranging from 0.5 to 0.95 along with mAP at 0.5 and 0.75 IoUs are reported.

\subsection{Implementation Details}
\label{sect:impl}

We implement our approach in PyTorch~\cite{paszke2017automatic}. For the baseline detectors, the implementations in the MMDetection~\cite{mmdetection} toolbox are adopted. For each detector, most training hyper-parameters are kept unchanged from the original settings used in the paper. The only setting we change in our experiments is the number of total training epochs. In the experiments of ablation studies, we train the models for 19 epochs and reduce the learning rate twice by a factor of 10 at the 16-th and 18-th epochs. In the experiments of MLC (Table~\ref{tab:mlc-all}), all models are trained for 24 epochs, and the learning rate is reduced at the 18-th and 22-th epochs. Notably, we pre-train the detectors in the original setting, and the MLC training strategy is enabled after the 12-th epoch. The $\alpha$ in Eq.~\eqref{eq:W-cls} and Eq.~\eqref{eq:W-loc} are set to 0 for FCOS$^{-}$, 0.5 for RetinaNet and 2 for RPN. We use ResNet-50-FPN~\cite{he2016deep,lin2017feature} as the network backbone.

\subsection{Ablation Studies}

To verify the effectiveness of mutual learning and IoU rescoring, we conduct ablation studies based on FCOS$^-$.
Considering that the candidate boxes are ranked by confidence when going through NMS, ideally the IoU should be monotonically increasing with respect to the confidence. The monotonic relationship between two variables can be well captured by the Spearman rank correlation coefficient~\cite{c1987the}. Thus we propose to quantify the divergence with the Spearman rank correlation between the classification accuracy and the localization quality. The more the correlation approaches 1, the better the divergence problem gets alleviated.

\noindent \textbf{Impact of Mutual Learning.}
To reduce the divergence, we may alternatively align the predictions of the two branches by imposing an MSE loss. Table~\ref{tab:comp-ml} lists the AP gains brought by different mutual learning strategies. With the mutual labeling-based mutual learning, AP gets boosted by 3 points from 34.5\% to 37.5\%. Prediction alignment brings a limited AP gain. This is because the constraint of an exact matching between the two tasks is too strong, which limits the learning of classification and localization. Notably, the Spearman rank correlation-based metric correlates well the AP value.

\noindent \textbf{Impact of IoU Rescoring.}
To verify the effectiveness of IUR, we compare it with the centerness prediction in FCOS~\cite{tian2019fcos}, which aims to measure the localization quality by centerness. As shown in Table~\ref{tab:comp-iur}, IUR improves the baseline significantly from 34.5\% to 37.3\%, and outperforms centerness by 0.4\%. In terms of the divergence, centerness barely improves the correlation, while IUR improves it from 0.32 to 0.50.

We also compare IUR with the IoU-NMS strategy introduced in IoU-Net~\cite{jiang2018acquisition}. For a fair comparison, we feed the same IoU predictions learned by IUR to IoU-NMS. Specifically, we disable the product operation of classification score and IoU prediction and replace the standard NMS with IoU-NMS. Surprisingly, the AP drops drastically from 37.3\% to 32.6\%. The performance degradation by IoU-NMS is also observed (not shown in the paper) for other detectors like RetinaNet. Since IoU-NMS uses the predicted IoUs to rank the detections, the accuracy of IoU estimation is critical. That is, IoU-NMS works for high quality IoU predictions only, and the learned IoUs by our light-weight network may not be accurate enough. On the contrary, the simple IoU rescoring scheme is more robust, and consistently improves the performance of various detectors (see also Table~\ref{tab:fcos-retinanet} and \ref{tab:rpn}).

\begin{table}[t]
\small
\centering
\caption{Complementary AP gains brought by ML and IUR for two detectors.}
\label{tab:fcos-retinanet}
\begin{tabular}{l|c|c|c}
\hline
Method & ML & IUR & AP (\%) \\
\hline
\multirow{4}{*}{FCOS$^{-}$}
& {} & {} & 34.5 \\
& {$\checkmark$} & {} & 37.5 \\
&  & {$\checkmark$} & 37.3 \\
& {$\checkmark$} & {$\checkmark$} & \textbf{38.7} \\
\hline
\multirow{4}{*}{RetinaNet}
& {} & {} & 36.3 \\
& {$\checkmark$} & {} & 37.1 \\
&  & {$\checkmark$} & 37.4 \\
& {$\checkmark$} & {$\checkmark$} & \textbf{37.8} \\
\hline
\end{tabular}
\end{table}

\begin{table}[tb]
\small
\centering
\caption{ML and IUR improve the average recall of region proposals by RPN.}
\label{tab:rpn}
\begin{tabular}{l|c|c|ccc}
\hline
Method & ML & IUR & AR$^{100}$ & AR$^{300}$ & AR$^{1000}$  \\
\hline
\multirow{4}{*}{RPN}
& {} & {} & 42.6 & 51.3 & 57.2 \\
& {$\checkmark$} & {} & 48.5 & 54.6 & \textbf{58.6} \\
&  & {$\checkmark$} & 45.0 & 52.8 & 58.0 \\
& {$\checkmark$} & {$\checkmark$} & \textbf{48.8} & \textbf{55.1} & \textbf{58.6} \\
\hline
\end{tabular}
\end{table}

\noindent \textbf{Combining Mutual Labeling and IUR.}
To verify the generalizability of our approach, we apply it to three detectors, i.e. FCOS$^{-}$, RetinaNet and RPN. As shown in Table~\ref{tab:fcos-retinanet} and Table~\ref{tab:rpn}, both mutual labeling and IUR consistently improve the baseline models. And their combination (MLC) achieves the best performance, which indicates that they are complementary to each other. Notably, MLC improves the average recall of region proposals by RPN in all settings, up to 6.2\% for the AR$^{100}$ setting.


\subsection{Results}

\begin{table}[t]
\small
\setlength{\tabcolsep}{5pt}
\centering
\caption{Final results on the validation set of the COCO dataset. For a fair comparison, we report the reference performance of the baseline detectors from our reproduction.}
\label{tab:mlc-all}
\begin{tabular}{l|c|ccccc}
\hline
Method & AP & AP$_{50}$ & AP$_{75}$ & AP$_{S}$ & AP$_{M}$ & AP$_{L}$  \\
\hline
FCOS& 36.9 & 55.8 & 39.1 & 20.4 & 40.1 & 49.2 \\
FCOS$^{-}$ & 34.6 & 54.5 & 36.0 & 19.3 & 39.1 & 45.0 \\
FCOS$^{-}$ + MLC & \textbf{38.7} & \textbf{57.6} & \textbf{41.7} & \textbf{21.8} & \textbf{42.5} & \textbf{51.0}\\
\hline
RetinaNet & 36.4 & 56.3 & 38.7 & 19.3 & 39.9 & 48.9\\
RetinaNet + MLC & \textbf{38.1} & \textbf{56.8} & \textbf{41.5} & \textbf{21.3} & \textbf{41.5} & \textbf{50.3}\\
\hline
\end{tabular}
\end{table}

The detailed final results of MLC applied to FCOS and RetinaNet using ResNet-50-FPN as the backbone are shown in Table~\ref{tab:mlc-all}. The higher APs than Table~\ref{tab:fcos-retinanet} are due to longer training as described in Sect.~\ref{sect:impl}. MLC outperforms the original FCOS with centerness by 1.8\%. Notably, we improve AP of the FCOS$^{-}$ baseline by 4.1\%. For RetinaNet, we achieve an AP gain of 1.7\%.

The improvement brought by MLC varies for AP at different IoU thresholds. From Table~\ref{tab:mlc-all}, we can see MLC improves AP$_{75}$ more significantly than AP$_{50}$, e.g. 5.7\% versus 3.1\% for FCOS$^-$ and 2.8\% versus 0.5\% for RetinaNet. In other words, the localization quality of the detections has been greatly improved. This observation clearly confirms our motivation that by reducing the divergence between classification of localization, good detections are more likely to be selected and bad detections are more likely to be suppressed.

\section{Conclusion}

The mutual learning framework can be interpreted as a procedure of consensus decision-making between the classification and localization tasks. In the beginning, they barely agree with each other on deciding whether a candidate box is good or bad. With mutually set target labels, they are able to teach and learn from each other at the same time, and finally reach a consensus. In this way, the divergence gets reduced.

In summary, we systematically investigate the divergence between the localization and classification tasks of CNN-based detectors. To reduce the divergence, we introduce the concept of mutual learning into object detection, and propose the mutual labeling strategy. During inference, we introduce a simple IoU rescoring scheme, which complements mutual learning. The extensive experiments clearly validate the effectiveness and versatility of the proposed approach. Besides, we propose a Spearman rank correlation-based metric to quantify the degree of divergence for a given detector, which may help future research in this direction.

\bibliographystyle{IEEEbib}
\bibliography{main}

\begin{thebibliography}{10}

\bibitem{he2017mask}
Kaiming {He}, Georgia {Gkioxari}, Piotr {Dollar}, and Ross {Girshick},
\newblock ``Mask r-cnn,''
\newblock in {\em ICCV}, 2017, pp. 2980--2988.

\bibitem{papandreou2017towards}
George {Papandreou}, Tyler {Zhu}, Nori {Kanazawa}, Alexander {Toshev}, Jonathan
  {Tompson}, Chris {Bregler}, and Kevin {Murphy},
\newblock ``Towards accurate multi-person pose estimation in the wild,''
\newblock in {\em CVPR}, 2017, pp. 3711--3719.

\bibitem{liu2016ssd}
Wei {Liu}, Dragomir {Anguelov}, Dumitru {Erhan}, Christian {Szegedy}, Scott~E.
  {Reed}, Cheng-Yang {Fu}, and Alexander~C. {Berg},
\newblock ``Ssd: Single shot multibox detector,''
\newblock in {\em ECCV}, 2016, pp. 21--37.

\bibitem{ren2017faster}
Shaoqing Ren, Kaiming He, Ross~B Girshick, and Jian Sun,
\newblock ``Faster r-cnn: Towards real-time object detection with region
  proposal networks,''
\newblock {\em IEEE Transactions on Pattern Analysis and Machine Intelligence},
  vol. 39, no. 6, pp. 1137--1149, 2017.

\bibitem{lin2017focal}
Tsung-Yi {Lin}, Priya {Goyal}, Ross~B. {Girshick}, Kaiming {He}, and Piotr
  {Dollár},
\newblock ``Focal loss for dense object detection,''
\newblock in {\em ICCV}, 2017, pp. 2999--3007.

\bibitem{yu2016unitbox}
Jiahui {Yu}, Yuning {Jiang}, Zhangyang {Wang}, Zhimin {Cao}, and Thomas~S.
  {Huang},
\newblock ``Unitbox: An advanced object detection network,''
\newblock in {\em Proceedings of the 24th ACM international conference on
  Multimedia}, 2016, pp. 516--520.

\bibitem{tian2019fcos}
Zhi {Tian}, Chunhua {Shen}, Hao {Chen}, and Tong {He},
\newblock ``Fcos: Fully convolutional one-stage object detection,''
\newblock in {\em ICCV}, 2019, pp. 9627--9636.

\bibitem{jiang2018acquisition}
Borui {Jiang}, Ruixuan {Luo}, Jiayuan {Mao}, Tete {Xiao}, and Yuning {Jiang},
\newblock ``Acquisition of localization confidence for accurate object
  detection,''
\newblock in {\em ECCV}, 2018, pp. 816--832.

\bibitem{zhang2018deep}
Ying {Zhang}, Tao {Xiang}, Timothy~M. {Hospedales}, and Huchuan {Lu},
\newblock ``Deep mutual learning,''
\newblock in {\em CVPR}, 2018, pp. 4320--4328.

\bibitem{lin2014microsoft}
Tsung-Yi {Lin}, Michael {Maire}, Serge~J. {Belongie}, James {Hays}, Pietro
  {Perona}, Deva {Ramanan}, Piotr {Dollár}, and C.~Lawrence {Zitnick},
\newblock ``Microsoft coco: Common objects in context,''
\newblock in {\em ECCV}, 2014, pp. 740--755.

\bibitem{redmon2016you}
Joseph {Redmon}, Santosh~Kumar {Divvala}, Ross~B. {Girshick}, and Ali
  {Farhadi},
\newblock ``You only look once: Unified, real-time object detection,''
\newblock in {\em CVPR}, 2016, pp. 779--788.

\bibitem{redmon2017yolo9000}
Joseph {Redmon} and Ali {Farhadi},
\newblock ``Yolo9000: Better, faster, stronger,''
\newblock in {\em CVPR}, 2017, pp. 6517--6525.

\bibitem{redmon2018yolov3}
Joseph {Redmon} and Ali {Farhadi},
\newblock ``Yolov3: An incremental improvement.,''
\newblock {\em arXiv preprint arXiv:1804.02767}, 2018.

\bibitem{girshick2014rich}
Ross {Girshick}, Jeff {Donahue}, Trevor {Darrell}, and Jitendra {Malik},
\newblock ``Rich feature hierarchies for accurate object detection and semantic
  segmentation,''
\newblock in {\em CVPR}, 2014, pp. 580--587.

\bibitem{girshick2015fast}
Ross {Girshick},
\newblock ``Fast r-cnn,''
\newblock in {\em ICCV}, 2015, pp. 1440--1448.

\bibitem{zhong2020}
Qiaoyong Zhong, Chao Li, Yingying Zhang, Di~Xie, Shicai Yang, and Shiliang Pu,
\newblock ``Cascade region proposal and global context for deep object
  detection,''
\newblock {\em Neurocomputing}, vol. 395, pp. 170--177, 2020.

\bibitem{lin2017feature}
Tsung-Yi {Lin}, Piotr {Dollár}, Ross~B. {Girshick}, Kaiming {He}, Bharath
  {Hariharan}, and Serge~J. {Belongie},
\newblock ``Feature pyramid networks for object detection,''
\newblock in {\em CVPR}, 2017, pp. 936--944.

\bibitem{huang2015densebox}
Lichao {Huang}, Yi~{Yang}, Yafeng {Deng}, and Yinan {Yu},
\newblock ``Densebox: Unifying landmark localization with end to end object
  detection.,''
\newblock {\em arXiv preprint arXiv:1509.04874}, 2015.

\bibitem{zhang2020bridging}
Shifeng {Zhang}, Cheng {Chi}, Yongqiang {Yao}, Zhen {Lei}, and Stan~Z. {Li},
\newblock ``Bridging the gap between anchor-based and anchor-free detection via
  adaptive training sample selection,''
\newblock in {\em CVPR}, 2020, pp. 9759--9768.

\bibitem{ke2020multiple}
Wei {Ke}, Tianliang {Zhang}, Zeyi {Huang}, Qixiang {Ye}, Jianzhuang {Liu}, and
  Dong {Huang},
\newblock ``Multiple anchor learning for visual object detection,''
\newblock in {\em CVPR}, 2020, pp. 10206--10215.

\bibitem{kim2020probabilistic}
Kang {Kim} and Hee~Seok {Lee},
\newblock ``Probabilistic anchor assignment with iou prediction for object
  detection.,''
\newblock in {\em ECCV}, 2020.

\bibitem{zhu2020autoassign}
Benjin {Zhu}, Jianfeng {Wang}, Zhengkai {Jiang}, Fuhang {Zong}, Songtao {Liu},
  Zeming {Li}, and Jian {Sun},
\newblock ``Autoassign: Differentiable label assignment for dense object
  detection.,''
\newblock {\em arXiv preprint arXiv:2007.03496}, 2020.

\bibitem{otsu1979a}
Nobuyuki {Otsu},
\newblock ``A threshold selection method from gray-level histograms,''
\newblock {\em IEEE Transactions on Systems, Man, and Cybernetics}, vol. 9, no.
  1, pp. 62--66, 1979.

\bibitem{paszke2017automatic}
Adam Paszke, Sam Gross, Soumith Chintala, Gregory Chanan, Edward Yang, Zachary
  DeVito, Zeming Lin, Alban Desmaison, Luca Antiga, and Adam Lerer,
\newblock ``Automatic differentiation in {PyTorch},''
\newblock in {\em NIPS Autodiff Workshop}, 2017.

\bibitem{mmdetection}
Kai Chen, Jiaqi Wang, Jiangmiao Pang, Yuhang Cao, Yu~Xiong, Xiaoxiao Li,
  Shuyang Sun, Wansen Feng, Ziwei Liu, Jiarui Xu, Zheng Zhang, Dazhi Cheng,
  Chenchen Zhu, Tianheng Cheng, Qijie Zhao, Buyu Li, Xin Lu, Rui Zhu, Yue Wu,
  Jifeng Dai, Jingdong Wang, Jianping Shi, Wanli Ouyang, Chen~Change Loy, and
  Dahua Lin,
\newblock ``{MMDetection}: Open mmlab detection toolbox and benchmark,''
\newblock {\em arXiv preprint arXiv:1906.07155}, 2019.

\bibitem{he2016deep}
Kaiming {He}, Xiangyu {Zhang}, Shaoqing {Ren}, and Jian {Sun},
\newblock ``Deep residual learning for image recognition,''
\newblock in {\em CVPR}, 2016, pp. 770--778.

\bibitem{c1987the}
Spearman {C},
\newblock ``The proof and measurement of association between two things. by c.
  spearman, 1904.,''
\newblock {\em American Journal of Psychology}, vol. 100, pp. 441, 1987.

\end{thebibliography}

\end{document}